\documentclass[lettersize,journal]{IEEEtran}
\usepackage{amsmath,amsfonts}
\usepackage{algorithmic}
\usepackage{algorithm}
\usepackage{array}
\usepackage[caption=false,font=normalsize,labelfont=sf,textfont=sf]{subfig}
\usepackage{textcomp}
\usepackage{stfloats}
\usepackage{verbatim}
\usepackage{graphicx}
\usepackage{xcolor}
\usepackage{booktabs} 
\usepackage{amsmath}  

\usepackage[hyphens]{url}
\usepackage[numbers]{natbib}
\usepackage[switch,columnwise]{lineno}

\usepackage{hyperref}

\begin{document}
\author{Jianman Lin, Tianshui Chen, Chunmei Qing, Zhijing Yang, Shuangping Huang, Yuheng Ren, Liang Lin \emph{Fellow, IEEE}
\thanks{Jianman Lin, Chunmei Qing, and Shuangping Huang are with the South China University of Technology (Emails: linjianmancjx@gmail.com, qchm@scut.edu.cn, eehsp@scut.edu.cn). Tianshui Chen and Zhijing Yang are with the Guangdong University of Technology (Emails: tianshuichen@gmail.com, yzhj@gdut.edu.cn). Yuheng Ren is with Jimei University and the Xiamen Kunlu AI Research Institute (Email: szcyxy@jmu.edu.cn). Liang Lin is with Sun Yat-sen University (Email: linliang@ieee.org).}}

\title{Neural Scene Designer: Self-Styled Semantic Image Manipulation}




\maketitle
\begin{abstract}
Maintaining stylistic consistency is crucial for the cohesion and aesthetic appeal of images, a fundamental requirement in effective image editing and inpainting. However, existing methods primarily focus on the semantic control of generated content, often neglecting the critical task of preserving this consistency.
In this work, we introduce the Neural Scene Designer (NSD), a novel framework that enables photo-realistic manipulation of user-specified scene regions while ensuring both semantic alignment with user intent and stylistic consistency with the surrounding environment. NSD leverages an advanced diffusion model, incorporating two parallel cross-attention mechanisms that separately process text and style information to achieve the dual objectives of semantic control and style consistency.
To capture fine-grained style representations, we propose the Progressive Self-style Representational Learning (PSRL) module. This module is predicated on the intuitive premise that different regions within a single image share a consistent style, whereas regions from different images exhibit distinct styles. The PSRL module employs a style contrastive loss that encourages high similarity between representations from the same image while enforcing dissimilarity between those from different images.
Furthermore, to address the lack of standardized evaluation protocols for this task, we establish a comprehensive benchmark. This benchmark includes competing algorithms, dedicated style-related metrics, and diverse datasets and settings to facilitate fair comparisons. Extensive experiments conducted on our benchmark demonstrate the effectiveness of the proposed framework.

\end{abstract}

\section{Introduction}
\begin{figure}[!t]
    \centering
    \includegraphics[width=0.47\textwidth]{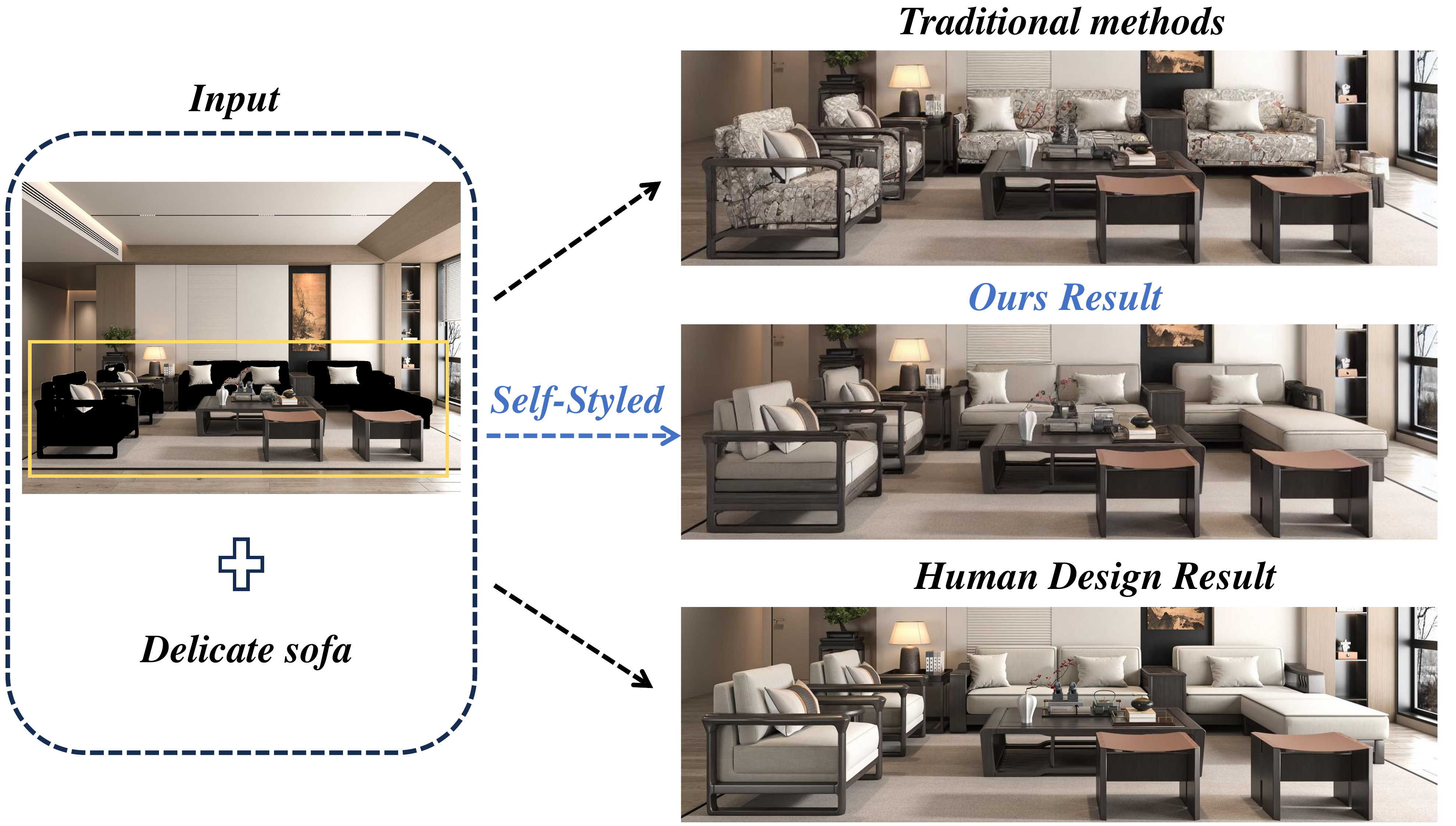}
    \caption{Illustration of traditional image editing\&inpainting and the proposed NSD results. NSD ensures that content generated for specified regions is not only semantically coherent with human intent but also seamlessly harmonized in style with its surrounding environment.}
    \label{fig:baseline-comparison}
\end{figure}

Stylistic consistency is a hallmark of professionally designed visual scenes, playing a crucial role in enhancing their overall aesthetic appeal and cohesion. This principle is particularly evident in domains like interior design, where professionals meticulously adjust colors, textures, and objects to achieve a harmonious and attractive overall style. While modern image editing and inpainting techniques aim to manipulate or replace specific regions to align with user intent, their primary focus has been on semantic control---ensuring the generated content matches the user's description. Consequently, the critical aspect of maintaining stylistic consistency with the surrounding, unedited regions is often overlooked (see Fig.~\ref{fig:baseline-comparison}). To bridge this gap, we introduce and address the task of Self-Styled Semantic Image Manipulation (S3IM). The goal of S3IM is to generate content that is not only semantically aligned with user instructions but also stylistically consistent with the rest of the image.

Traditional image inpainting techniques primarily focus on filling masked regions by propagating information from the surrounding context~\cite{rombach2022high, wang2023imagen, ju2024brushnet, zhuang2023task, yu2018generative, pathak2016context, lugmayr2022repaint, antonio2004region}. However, these methods often fail to synthesize content that is coherent both semantically and stylistically, largely due to their lack of textual guidance and explicit style modeling. More recent text-guided inpainting methods~\cite{rombach2022high, wang2023imagen, ju2024brushnet, zhuang2023task} have addressed the semantic gap by leveraging textual descriptions to guide content generation. These models also implicitly attempt to maintain stylistic consistency by capitalizing on the self-attention mechanisms within pre-trained diffusion models. Despite improving semantic alignment, they still struggle to achieve robust stylistic coherence. This limitation stems from the inherent training bias of large-scale diffusion models, which are typically trained on vast image-text datasets that prioritize semantic correspondence over fine-grained stylistic details.

Explicitly modeling the style representation of a scene image is crucial for addressing the S3IM task. 
One line of research~\cite{liu2023stylecrafter, ye2023ip, wang2024instantstyle} treats style as a form of semantic information, employing visual language models (VLMs)~\cite{radford2021learning, kwon2022masked} with carefully designed prompts to extract stylistic attributes. 
However, the efficacy of this approach is limited, as style is predominantly influenced by low-level cues like color and texture, which do not consistently align with the high-level semantic information captured by VLMs. 
An alternative approach~\cite{zhang2022domain, yang2024emogen} leverages style categories (e.g., minimalist or European styles for indoor scenes) as supervision to learn style representations. 
Yet, this category-level supervision often proves too coarse, failing to capture the nuanced stylistic variations unique to each image.

In this work, we introduce the Neural Scene Designer (NSD), which learns fine-grained style representations and integrates them with semantic descriptions in advanced diffusion models \cite{ju2024brushnet} to ensure both semantic alignment and style consistency. To capture subtle style variations, we introduce a Progressive Self-style Representational Learning (PSRL) module, which operates on the reasonable assumption that different regions within the same scene image maintain a consistent style, while regions across different images exhibit distinct styles. This module allows for semantic-independent and instance-level supervision that discards irrelevant semantic information while capturing sophisticated style details. Specifically, PSRL employs a style contrastive loss to ensure that the style representations within the same image are highly similar, whereas those from different images are markedly different. Two parallel cross-attention mechanisms integrate the learned style representation with text embeddings derived from semantic descriptions into the diffusion models. Additionally, a reference network then aggregates information from unmasked regions and incorporates it into the diffusion denoising process to maintain content coherence in the unmasked regions.

The advancement of S3IM has been significantly hindered by the absence of a standardized evaluation benchmark. To address this critical gap and foster future research, we establish the first unified benchmark for this task. A key deficiency in current evaluation practices is the lack of metrics specifically designed to assess stylistic consistency, as most protocols prioritize semantic alignment. We bridge this gap by incorporating three powerful metrics suited for this purpose: Cosine Style Distance (CSD)~\cite{somepalli2024measuring}, Human Preference Score (HPS)~\cite{wu2023human}, and ImageReward (IR)~\cite{xu2024imagereward}. Furthermore, while existing methods are typically evaluated on general-purpose datasets, we argue that professionally designed indoor scenes, with their inherent stylistic coherence, provide a more suitable testbed for S3IM. Consequently, we introduce \textbf{S3IMIndoorData}, a new dataset specifically curated for evaluating stylistic consistency in indoor scene manipulation. Under this comprehensive benchmark---comprising tailored metrics and a specialized dataset---we conduct extensive and fair comparisons against several state-of-the-art algorithms across diverse experimental settings.

The contributions of this work are fourfold. 
First, we propose the Neural Scene Designer (NSD) framework, a novel approach for completing missing regions or objects with both semantic alignment to human intent and stylistic consistency with the surrounding environment. 
To our knowledge, NSD is the first framework designed to explicitly model and preserve stylistic consistency for this task. 
Second, we design the Progressive Self-style Representational Learning (PSRL) module, which leverages semantic-independent, instance-level supervision to learn robust and fine-grained style representations. 
Third, we establish a fair and comprehensive benchmark for S3IM to systematically evaluate leading methods and foster future research in this area. 
Finally, we follow the unified benchmark to conduct extensive experiments to verify the effectiveness of the proposed algorithm. Our results demonstrate that the NSD framework achieves superior performance in terms of style consistency, text alignment, image quality, and preservation of masked regions. All codes, trained models, and datasets are available at \url{https://github.com/jianmanlincjx/NSD.git}.

\section{Related Work}
In this section, we provide a detailed introduction to traditional image inpainting, text-guided image inpainting, and the conceptually related field of style representation, to establish the context for our proposed approach.
\subsection{Traditional Image Inpainting}
Traditional image inpainting techniques focus on synthesizing plausible content for masked regions by propagating information from the surrounding context. The advent of deep learning, particularly Generative Adversarial Networks (GANs)~\cite{liu2021pd, zhang2022gan, zhao2021large, wang2021dynamic, quan2022image, li2023transformer, chen2024heterogeneous, chen2024dynamic}, has led to significant progress in this area. Early deep learning approaches focused on integrating structural and latent priors to guide content generation~\cite{lahiri2020prior}. Subsequent works introduced multi-stage refinement strategies, employing networks with varying receptive fields to handle missing regions of different scales~\cite{zhao2021large}. To better leverage information from uncorrupted areas, DSNet~\cite{yu2022high} proposed dynamically selecting valid pixels and normalization methods to enhance detail. Addressing the challenge of large masks, LaMa~\cite{suvorov2022resolution} employed fast Fourier convolutions, while CM-GAN~\cite{zheng2022cmgan} utilized global-spatial modulation blocks to capture long-range dependencies. More recently, methods have incorporated context-aware mechanisms, such as the adaptive modules in CANet~\cite{deng2023context} and the mask-aware dynamic filtering in~\cite{zhu2021image}, to achieve more refined results.

Despite their success in generating coherent textures and structures, these context-based models share a fundamental limitation: they lack mechanisms for explicit semantic control and stylistic guidance. Operating without textual prompts or explicit style modeling, the synthesized content, while plausible, often fails to align with specific user intent or maintain stylistic consistency with the broader scene.

\subsection{Text-Guided Image Inpainting}
To overcome the semantic limitations of traditional methods, text-guided image inpainting leverages natural language descriptions to direct content generation within masked regions. This paradigm has been revolutionized by the emergence of diffusion models~\cite{song2020denoising, ho2022classifier, corneanu2024latentpaint, xie2023smartbrush, yang2023uni, yu2023inpaint, manukyan2023hd, lin2025geometry}, which excel at synthesizing high-fidelity, semantically rich content. A primary challenge in this domain is effectively conditioning the generation process on the unmasked context. One line of work achieves this by modifying the sampling process of a pre-trained model. For example, Blended Latent Diffusion~\cite{avrahami2023blended} composes the latent representation at each denoising step by combining the model's prediction for the masked region with the known latent from the original image. Similarly, other approaches~\cite{nichol2021improved, corneanu2024latentpaint} perform a forward-backward fusion in the latent space to propagate context, enabling efficient inpainting without costly retraining; LatentPaint~\cite{corneanu2024latentpaint} further enhances this with a novel propagation module. Another prominent strategy involves adapting the model's architecture. A straightforward approach is to fine-tune a model specifically for inpainting, as done in Stable Diffusion Inpainting~\cite{rombach2022high}, which feeds the masked image and mask as additional inputs to the U-Net. More sophisticated methods introduce dedicated modules for context injection. BrushNet~\cite{ju2024brushnet}, for instance, employs a separate encoding branch for the unmasked image, allowing its features to be controllably integrated into a frozen text-to-image model. MMGInpainting~\cite{zhang2024mmginpainting} designs a Semantic Fusion Encoder to better merge visual context with textual guidance. Beyond architectural modifications, researchers have also explored novel training frameworks, such as the causal-based approach of CaPaint~\cite{NEURIPS2024_c2d82a42} for improved reasoning, and unified models like PowerPoint~\cite{zhuang2023task} that handle multiple editing tasks via task-specific prompts.

While existing models excel at semantic control, they often fail to maintain stylistic coherence—a critical challenge that has spurred recent research in style-aware generation. A prominent line of research, including works like StyleCrafter \cite{liu2023stylecrafter}, InstantStyle \cite{wang2024instantstyle}, and InstantStyle-Plus \cite{wang2024instantstyle-plus}, focuses on applying style from an external reference image. These methods employ sophisticated techniques, from dedicated adapters to content-preserving modules like ControlNet, to manage the process of external style transfer. This paradigm, however, is fundamentally different from the problem we address: preserving an image's own inherent style during local editing. To this end, our NSD framework leverages the PSRL module, a component specifically engineered to learn this internal style embedding. This ensures that generated content is harmonized not only semantically with the text prompt but also stylistically with the surrounding scene.

\begin{figure*}[htp]
  \centering
  \includegraphics[width=0.95\textwidth]{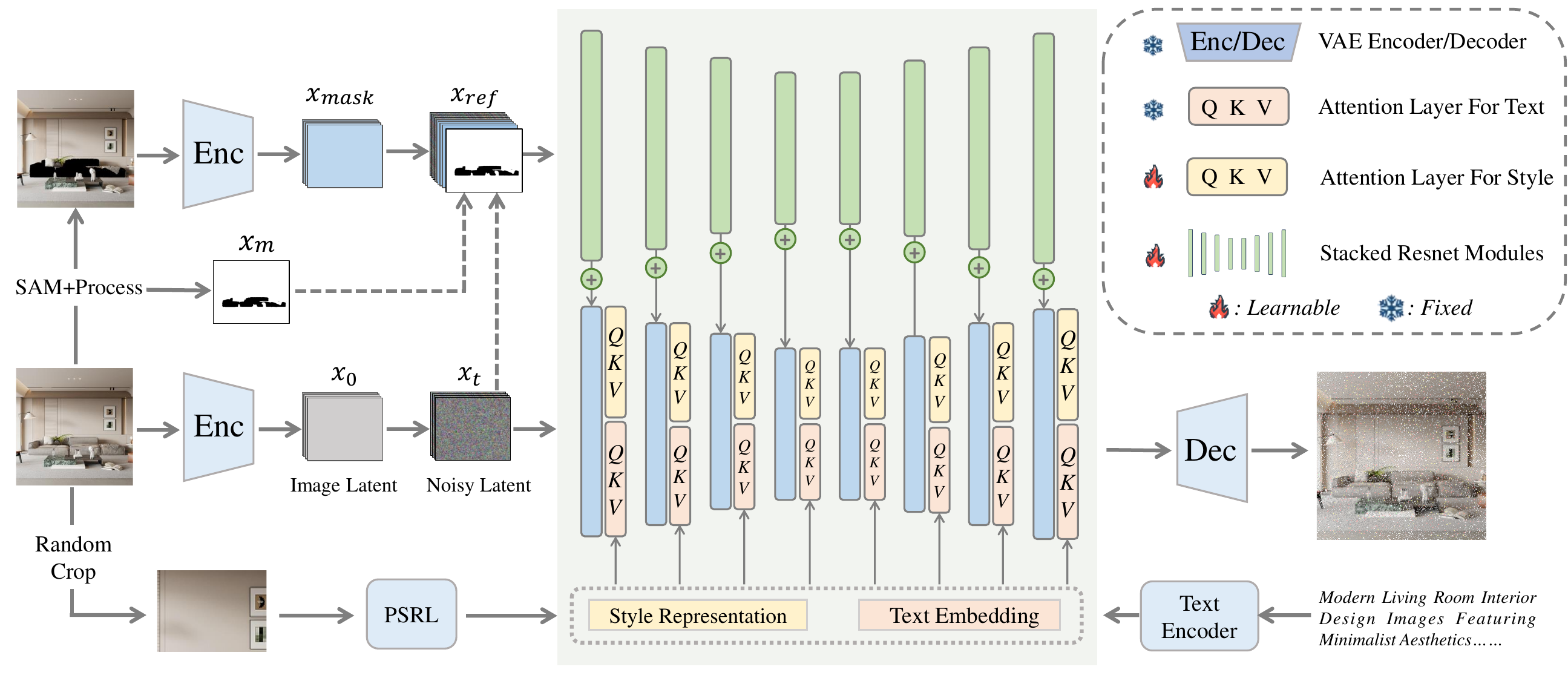} 
  \caption{The training process of the proposed framework is as follows (the inference process is similar but involves iterative denoising): It first utilizes the learned Progressive Self-style Representational Learning (PSRL) Module to accurately capture fine-grained style representations from the scene. Then, two parallel cross-attention mechanisms embed both style and semantic information into the diffusion model. Meanwhile, a reference network aggregates information from the unmasked regions and integrates it into the diffusion denoising process.}
  \label{fig: framework}
\end{figure*}

\section{Method}
Neural Scene Designer (NSD) builds upon an advanced diffusion model, incorporating two parallel cross-attention mechanisms to seamlessly integrate both textual and stylistic information. This ensures semantic alignment with user intent while maintaining stylistic consistency with the surrounding environment. Additionally, NSD utilizes Progressive Self-style Representational Learning (PSRL) to capture inherent human preferences, enabling the model to learn precise and fine-grained style representations. An overview of the NSD framework is presented in Fig. \ref{fig: framework}.

\subsection{Neural Scene Designer}

The proposed Neural Scene Designer (NSD) is built upon a pre-trained diffusion model to generate high-quality images. 
To ensure the generated content aligns with both user-provided semantic intent and the scene's intrinsic style, NSD incorporates two parallel cross-attention mechanisms that separately inject textual and stylistic features into the denoising U-Net. 
Furthermore, it utilizes a dedicated reference network to integrate information from unmasked regions, ensuring their high-fidelity preservation.

Formally, given an input image, a pre-trained autoencoder first maps it to a latent representation \(x_0\). 
A forward diffusion process then progressively adds Gaussian noise to \(x_0\) over \(T\) timesteps, governed by a Markov chain. The state at step \(t\) is defined as:
\begin{align}
\begin{split}
x_t = \alpha_t x_0 + \sigma_t \epsilon
\end{split}
\end{align}
where \( \alpha_t \) and \( \sigma_t \) are schedule-dependent functions controlling the signal and noise levels, respectively.

To guide the denoising process, we condition the model on both semantic and stylistic information. 
The semantic representation \( f_{sem} \) is extracted from the input text using a pre-trained CLIP text encoder~\cite{radford2021learning}, which is renowned for its effectiveness in capturing rich semantic details~\cite{ye2023ip}. 
For style, however, existing models lack the ability to extract accurate and fine-grained representations. 
We address this gap by designing a pre-trained \textbf{P}rogressive \textbf{S}elf-style \textbf{R}epresentational \textbf{L}earning (PSRL) module (detailed in Sec.~\ref{sec:psrl}) to derive the style representation \( f_{sty} \) from the unmasked image regions.

Following standard attention mechanisms~\cite{vaswani2017attention, xu2024self, xu2025exploiting}, \( f_{sem} \) and \( f_{sty} \) serve as keys and values, while the noisy latent \( x_t \) acts as the query. 
These are processed through two parallel cross-attention blocks to produce style-aware and semantic-aware features:
\begin{align}
\begin{split}
z_{sty}& = \text{Cross\_Att}(x_t, f_{sty}, f_{sty}) \\
z_{sem}& = \text{Cross\_Att}(x_t, f_{sem}, f_{sem})
\end{split}
\end{align}
these features are then fused to form the final conditional representation:
\begin{align}
\begin{split}
z = z_{sem} + \lambda \cdot z_{sty}
\end{split}
\label{eq:latent_fusion}
\end{align}
here, \( \lambda \) is a hyperparameter balancing the influence of semantic and style guidance. Throughout our experiments, we set $\lambda=1$. This choice provides an equal weighting to both conditioning signals, ensuring a balanced fusion of semantic correctness and stylistic consistency without requiring manual tuning.
The combined representation \( z \) thus encodes both the fine-grained style from unmasked regions and the semantic intent from the text description. The model is trained with the following objective:
\begin{align}
\begin{split}
L = \mathbb{E}_{x_0, \epsilon \sim \mathcal{N}(0, I), c, t} \left[ \left\| \epsilon - \epsilon_\theta \left( x_t, c, t \right) \right\|^2 \right]
\end{split}
\label{eq:simple_loss_NSD}
\end{align}
where the condition \( c \) comprises both style features $f_{sty}$ and semantic features $f_{sem}$. 
By minimizing the discrepancy between the ground-truth noise \( \epsilon \) and the predicted noise \( \epsilon_\theta \), the model's parameters are optimized via backpropagation.

\begin{figure}[!t]
    \centering
    \includegraphics[width=0.45\textwidth]{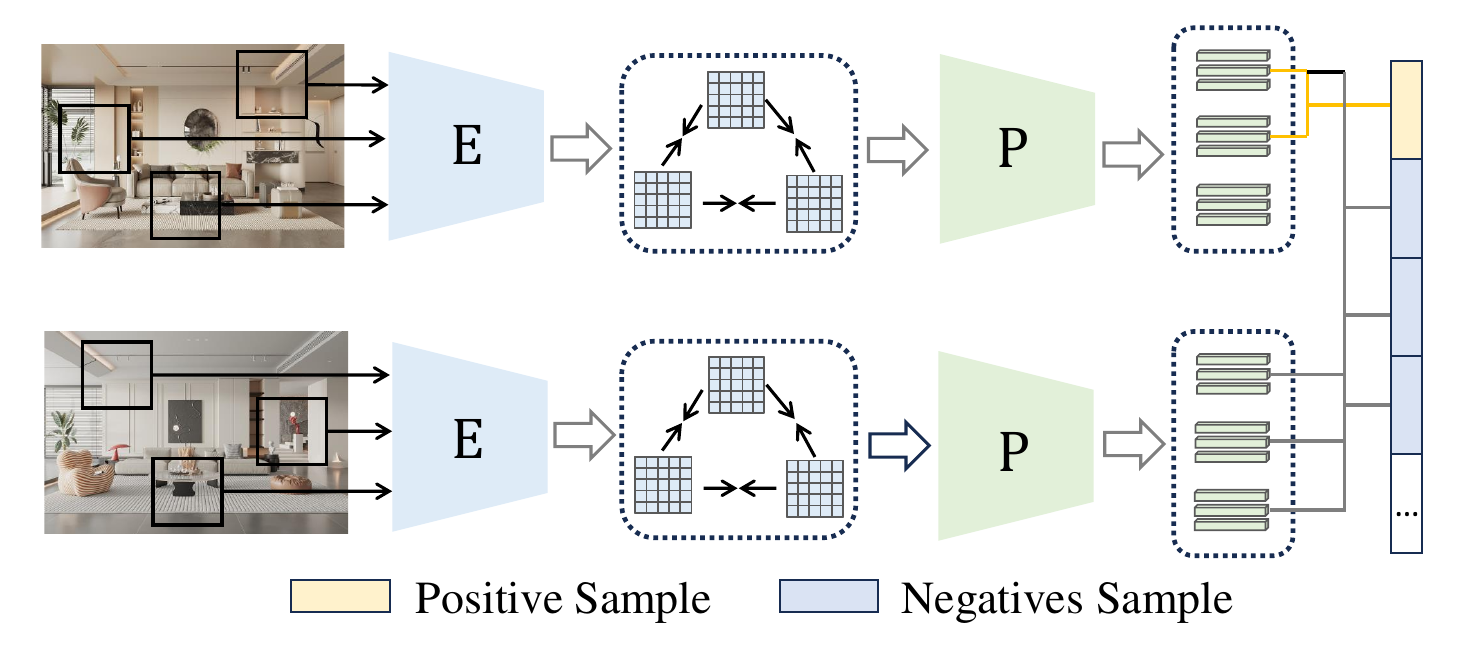}
    \caption{An illustration of Progressive Self-style Representational Learning. The process begins by pre-training a VGG-based network to extract foundational style features. Subsequently, self-style contrastive learning is employed to refine these features, yielding more accurate and fine-grained style representations.}
    \label{fig: PSRL Module}
\end{figure}

To preserve the known content of unmasked regions with high fidelity, we employ a reference network, \( \epsilon_{ref}(\cdot) \), rather than alternatives like ControlNet~\cite{zhang2023adding} that are optimized for sparse conditioning signals. Our reference network mirrors the architecture of the pre-trained U-Net but omits the text-conditioning cross-attention layers. This design compels it to function as a pure spatial feature extractor, ensuring the background is reconstructed faithfully. 
Specifically, \( \epsilon_{ref}(\cdot) \) takes as input the noisy latent \( x_t \), the masked image latent \( x_{mask} \), and the mask itself \( x_m \). These are resized to match dimensions and concatenated into a single tensor:
\begin{align}
\begin{split}
x_{ref} = \text{concat}(x_t, x_{mask}, x_m)
\end{split}
\end{align}
This tensor \( x_{ref} \) is then injected into the frozen diffusion model's layers via the reference network, enabling hierarchical, dense, per-pixel control over the generation process:
\begin{align}
\begin{split}
\epsilon_\theta \left( x_t, t, c \right)_i = \epsilon_\theta \left( x_t, t, c \right)_i +  \epsilon_{ref} \left( x_{ref}, t \right)_i
\end{split}
\end{align}
where \( i \) denotes the feature map from the $i$-th layer. As in ControlNet~\cite{zhang2023adding}, we use zero-convolution layers to connect the trainable reference network to the frozen U-Net backbone.

\subsection{Progressive Self-style Representational Learning}\label{sec:psrl}
In this section, we provide a systematic explanation of PSRL, including assumption validation, methodological details, and an explanation of why PSRL can extract more accurate and fine-grained style representations compared to existing methods.

\begin{figure}[!t]
    \centering
    \includegraphics[width=0.48\textwidth]{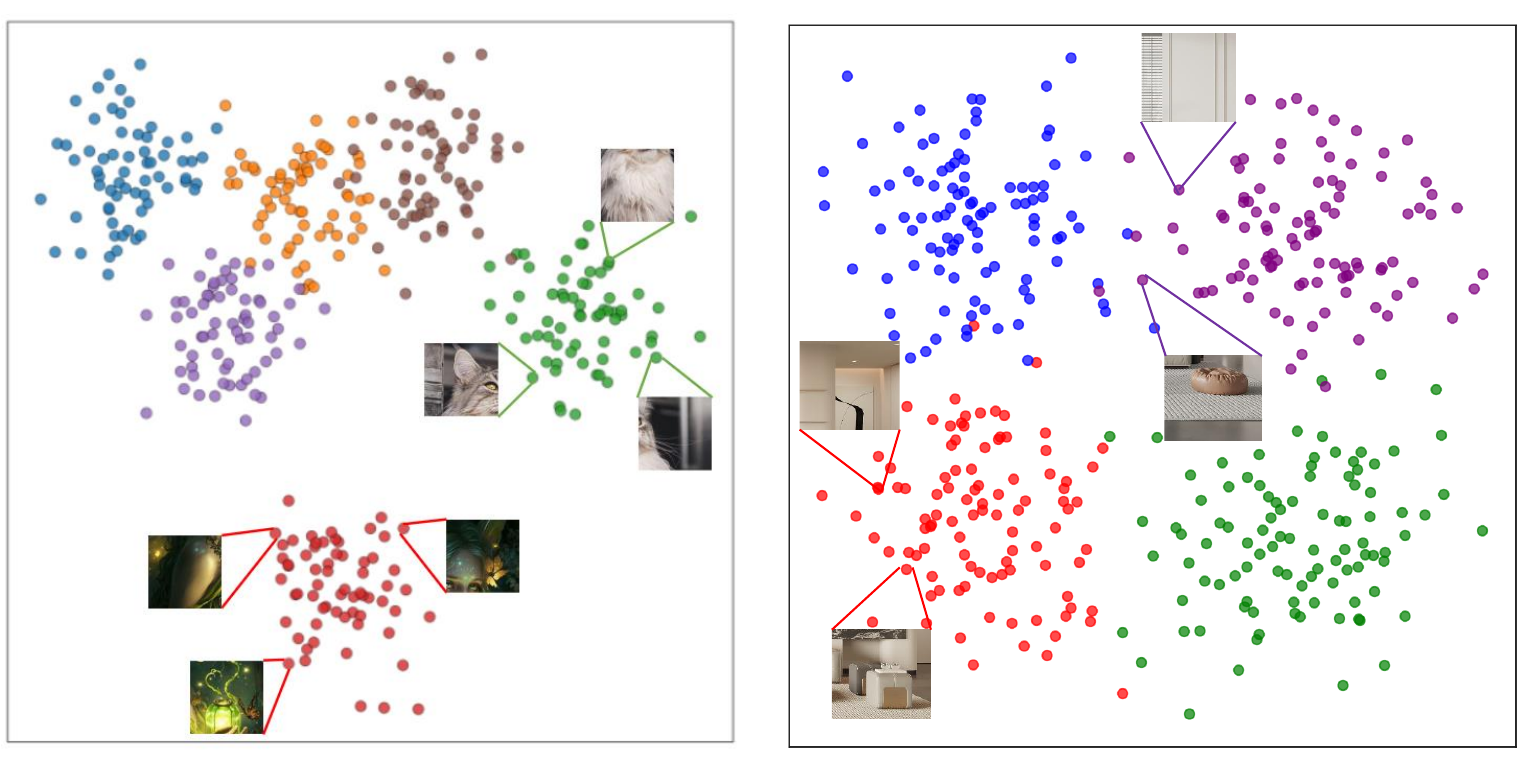}
    \caption{Dimensionality reduction visualization of general scene (left), together with the corresponding visualization of indoor scene (right): different colors represent different scenes, and different points within the same color represent different regions within the same scene. The image used is sourced from BrushBench \cite{ju2024brushnet}.}
    \label{fig:general_scene}
\end{figure}

\noindent\textbf{Assumption Validation. } PSRL is built upon the assumption that patches within the same scene exhibit stylistic consistency, while patches from different images display distinct styles. 
We empirically validate this assumption for both general and indoor scenes via a dimensionality reduction analysis. 
Specifically, we first crop a scene image into multiple non-overlapping patches and then employ a pre-trained style feature extractor~\cite{wu2022ccpl}, capable of handling diverse scenes from artistic to photo-realistic, to process these patches. 
The resulting features from the final layer are then visualized after dimensionality reduction. 
As shown in the left panel of Fig.~\ref{fig:general_scene}, patches from the same scene form a tight cluster, confirming that our assumption holds for general scenes. 
A similar phenomenon is observed for indoor scenes, as depicted in the right panel of Fig.~\ref{fig:general_scene}. 
This analysis provides strong evidence for the validity of our core assumption in both domains. 
Building on this insight, we designed the PSRL module to learn fine-grained and accurate style representations through contrastive learning.

\noindent\textbf{Method details. } Based on this assumption, we can apply contrastive learning, treating regions within the same image as having high style similarity while regions from different images are considered to have low style similarity to capture the style representation. However, scene images often contain multiple objects from diverse categories, making the direct application of contrastive learning challenging. This can hinder convergence and impede effective style learning due to semantic conflicts and ambiguity. As discussed in previous work \cite{wen2023cap, wu2022ccpl, zhang2022domain}, second-order statistics can mitigate semantic confusion and aligning these statistics can supervise learning style representations. However, second-order statistics struggle to capture color nuances and local textual distributions, which limits their applicability. To address this issue, PSRL exploits a progressive learning strategy, which first exploits second-order statistics to supervise learning initial style representation, followed by a learnable projector supervised by style contrastive learning to further extract final style information. We illustrate the progressive pipeline of PSRL in Fig. \ref{fig: PSRL Module}. 

Formally, given two scene images \( x \) and \( y \), we first randomly crop \( N \) regions from each image to obtain \( X_n \) and \( Y_n \), respectively. A VGG network \( E(\cdot) \) is then used to extract features from these cropped regions:

\begin{align}
\begin{split}
X_n^f &= \{E(X_n)\}_{n=1}^{N} \\
Y_n^f &= \{E(Y_n)\}_{n=1}^{N}
\end{split}
\end{align}
Following the approach in \cite{gatys2016image, wu2022ccpl, chen2021artistic}, we guide \( E(\cdot) \) to extract style features by constraining the second-order statistics of the features from any two elements within \( X_n^f \) or \( Y_n^f \) to be consistent. This is achieved by minimizing the following losses:

\begin{align}
\begin{split}
\mathcal{L}_{x} = \left\| \mu (X_n^{f,i}) - \mu (X_n^{f,j}) \right\| &+ \left\| \sigma (X_n^{f,i}) - \sigma (X_n^{f,j}) \right\| \\
\mathcal{L}_{y} = \left\| \mu (Y_n^{f,i}) - \mu (Y_n^{f,j}) \right\| &+ \left\| \sigma (Y_n^{f,i}) - \sigma (Y_n^{f,j}) \right\|
\end{split}
\end{align}
where \( i \) and \( j \) denote two non-overlapping regions within the same scene, and \( \mu \) and \( \sigma \) represent the mean and variance, respectively. To effectively extract and refine the final style representation, we propose using a learnable projector supervised by style contrastive learning \cite{chen2020simple, he2020momentum, chen2024learning, chen2025contrastive}. This is achieved by maximizing the feature similarity between any two elements within $X_n^f$ and minimizing the feature similarity between each element in $X_n^f$ and each element in $Y_n^f$:

\begin{equation}
\mathcal{L}_{xy} = -\log \textstyle \frac{\exp\left( \frac{P(X^{fi}_{n}) \cdot P(X^{fj}_{n})}{\tau} \right)}{\exp\left( \frac{P(X^{fi}_{n}) \cdot P(X^{fj}_{n})}{\tau} \right) + \sum_{k=1}^{N} \exp\left( \frac{P(X^{fi}_{n}) \cdot P(Y^{fk}_{n})}{\tau} \right)}
\end{equation}
where $P(\cdot)$ denotes a learnable projector and it is implemented using two stacked fully-connected layers cooperated with the rectified linear unit non-linear function. $\tau$ stands for a temperature hyper-parameter set to 0.07 by default. The final loss can be defined as:

\begin{equation}
\mathcal{L}_{psrl}= \mathcal{L}_{x} + \mathcal{L}_{y} + \mathcal{L}_{xy}
\label{eq:simple_loss_psrl}
\end{equation}
by minimizing losses through backpropagation, we can train the PSRL module to accurately extract fine-grained style representations under semantic-independent and instance-level supervision.

\begin{figure*}[!t]
    \centering
    \includegraphics[width=0.95\textwidth]{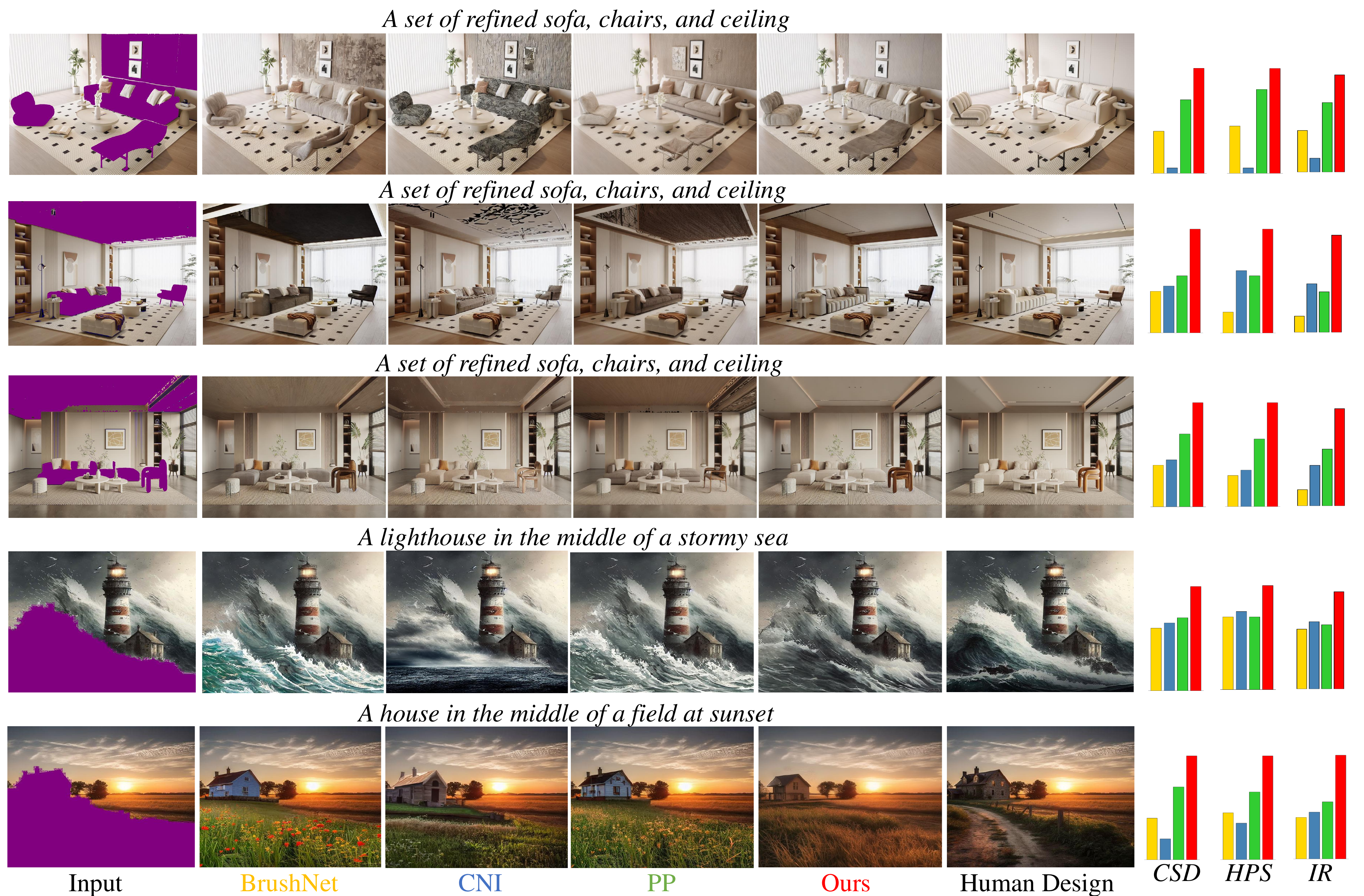}
    \caption{Qualitative comparisons of methods on the proposed S3IMIndoorData (first three rows) and BrushBench (last two rows) under the unfinetuned setting. (In the input image, the area to be generated is highlighted in purple.). \textit{Due to space limitations, we only selected BrushNet, CNI, and PP for visual comparison. We have supplemented more comprehensive visual comparisons in the supplementary materials.}}
    \label{fig:visual_baseline}
\end{figure*}

\noindent\textbf{Discussions. }
Current approaches for ensuring stylistic consistency fall into two main categories. The first leverages the implicit style priors within the self-attention layers of pre-trained diffusion models~\cite{chung2024style}. While this method can capture some stylistic relationships, it often suffers from an inherent bias towards semantics, frequently neglecting fine-grained stylistic details. The second category employs explicit, learnable style adapters~\cite{liu2023stylecrafter,wang2024instantstyle} to inject style features from an image prompt. However, the effectiveness of this approach is often capped by the limitations of the pre-trained encoders used for feature extraction. These encoders are typically designed for semantic tasks and thus fail to capture nuanced stylistic attributes, leading to suboptimal consistency.

PSRL, on the other hand, can capture accurate and fine-grained style representations. We explain this from two perspectives. First, PSRL naturally excludes semantic information from images. During the training process of PSRL, it constrains that the information extracted from different regions of a scene remains consistent. Intuitively, in a single scene, the information from different regions is naturally semantically distinct. Through backpropagation, PSRL automatically learns the shared information across different regions—style, thus allowing us to obtain an accurate style representation. Second, PSRL benefits from instance-level supervision. Compared to style representations learned under category-level supervision, PSRL learns style at the instance level, enabling it to capture subtle differences in style between images as well as maintain style consistency within different regions of the same image. Thus, we can obtain a fine-grained style representation.

\section{Evaluation Benchmark}
To ensure a fair and comprehensive comparison for the S3IM task, this section details our evaluation benchmark. We outline the competing algorithms, the dataset used, the evaluation metrics, and the experimental settings.

\subsection{Competing Algorithms}
We evaluate our proposed method against several state-of-the-art image manipulation techniques. The competing methods are as follows:

\begin{itemize}
    \item \textbf{Stable Diffusion-Inpainting (SDI)}~\cite{rombach2022high} (CVPR 2022): A fine-tuned version of Stable Diffusion that leverages object masks and image captions for text-guided image inpainting.
    
    \item \textbf{ControlNet-Inpainting (CNI)}~\cite{zhang2023adding} (ICCV 2023): Extends Stable Diffusion with a trainable branch that conditions the generation process on masked region information to guide image inpainting.
    
    \item \textbf{Blended Latent Diffusion (BLD)}~\cite{avrahami2023blended} (TOG 2023): A method that modifies the denoising process by sampling only the masked regions from a pre-trained diffusion model, while preserving the unmasked regions through a copy-paste mechanism in the latent space.
    
    \item \textbf{PowerPaint (PP)}~\cite{zhuang2023task} (ECCV 2024): A versatile framework that unifies multiple image manipulation tasks using learnable task prompts in conjunction with standard text prompts.
    
    \item \textbf{BrushNet}~\cite{ju2024brushnet} (ECCV 2024): Employs a dual-branch architecture to inject pixel-level masked image features into a pre-trained diffusion model, achieving coherent and high-quality image manipulation results.
\end{itemize}

\subsection{Dataset}
We utilize the \textbf{BrushData}~\cite{ju2024brushnet} dataset for training, which comprises 500,000 image samples, each paired with an automatically generated object mask and a corresponding caption. 
For evaluation, we use \textbf{BrushBench}~\cite{ju2024brushnet}, a benchmark containing 600 samples, each featuring a high-quality, human-annotated mask and a detailed text description. 
This training and evaluation setup is consistent with prior work. 
Furthermore, recognizing that human-designed indoor scenes often exhibit strong stylistic coherence, we constructed a new dataset tailored for this scenario, named \textbf{S3IMIndoorData}. 
This dataset contains approximately 40,000 training samples and 4,000 test samples. 
Each sample in S3IMIndoorData is annotated with an object mask and a corresponding textual description, making it highly suitable for S3IM evaluation.

\subsection{Metrics} 
The goal of S3IM is to generate content for designated regions or objects that is both semantically aligned with human intent and stylistically consistent with the surrounding environment. 
Accordingly, we evaluate performance from multiple perspectives. 
To assess stylistic consistency, we employ three metrics: \textbf{CSD}~\cite{somepalli2024measuring}, \textbf{HPS}~\cite{wu2023human}, and \textbf{IR}~\cite{xu2024imagereward}. 
CSD directly captures style features from the image and measures consistency by calculating the similarity between the generated patch and its surrounding context. 
IR and HPS are reward models trained on large-scale datasets to reflect human preferences; since stylistically coherent images are typically more visually pleasing, we leverage these models as effective proxies for style consistency. 
To evaluate semantic alignment, we use \textbf{CLIP Similarity (CLIP Sim)}~\cite{wu2021godiva} to measure the correspondence between the generated images and their text prompts. 
Finally, to assess overall generation quality, we utilize the \textbf{Aesthetic Score (AS)}~\cite{schuhmann2022laion} and compute the \textbf{Peak Signal-to-Noise Ratio (PSNR)} on the unmasked regions to quantify the preservation fidelity of the original content.

\subsection{Implementation Details}
This section details the implementation of each method under three distinct settings to ensure fair comparison and enable reproducibility.

In the first setting, designed to assess general performance, all models are trained on \textbf{BrushData}~\cite{ju2024brushnet} and evaluated on \textbf{BrushBench}~\cite{ju2024brushnet}. 
For our method, \textbf{NSD}, the PSRL module is first pre-trained on general scenes using an Adam optimizer with a batch size of 16 and a learning rate of $1 \times 10^{-4}$ on a single NVIDIA 4090 GPU. 
The pre-trained PSRL module is then integrated into the full NSD framework, which is subsequently trained for 500,000 steps on 8 NVIDIA V100 GPUs with an input image size of 512$\times$512. 
For all \textbf{other baselines}, we adopt their officially released checkpoints pre-trained on BrushData and evaluate them directly on BrushBench.

In the second setting, we evaluate the zero-shot generalization ability of the models. 
Specifically, we take the models trained on BrushData from the first setting and evaluate them directly on our \textbf{S3IMIndoorData} test set without any fine-tuning. 
For \textbf{NSD}, this means using the PSRL module that was pre-trained on general scenes.

In the third setting, we assess the performance of select models on indoor scenes by fine-tuning and testing them on \textbf{S3IMIndoorData}. 
For our \textbf{NSD}, the PSRL module is pre-trained from scratch on the S3IMIndoorData training set, after which the full model is fine-tuned for 20,000 steps. 
For \textbf{BrushNet}~\cite{ju2024brushnet}, we use its official implementation to fine-tune its pre-trained checkpoint on S3IMIndoorData for 20,000 steps. 
Similarly, for \textbf{CNI}~\cite{zhang2023adding}, we first preprocess the indoor data to extract Canny edge maps as conditions and then fine-tune the official pre-trained model for 20,000 steps.

\section{Experiments}
In this section, we provide a comprehensive assessment of the proposed NSD framework. This includes detailed implementation details, along with quantitative and qualitative analyses conducted based on the established evaluation benchmark. Additionally, we present insights from a user study and perform extensive ablation experiments to validate the effectiveness of each component within the NSD framework.
\subsection{Implementation details}

\noindent\textbf{Paired Data Construction.} This section details the construction process for the paired data used to train the PSRL module, outlining our approach for both indoor and general scenes. 

For the \textbf{indoor scene} setting, we scraped a collection of 35,000 high-quality interior design images and manually categorized them into 14 distinct style classes (e.g., Minimalist, Scandinavian, Industrial). 
During training, we form a negative pair by selecting two images from different style classes. 
From each of these two images, we then extract $N=10$ non-overlapping patches of size 128$\times$128 to serve as positive samples for contrastive learning. 
These parameters were carefully chosen: a 128$\times$128 patch size is large enough to capture salient style patterns (e.g., textures, color palettes) yet small enough to avoid excessive semantic diversity within a single patch. Using $N=10$ non-overlapping patches provides sufficient intra-image variation for robust learning while remaining feasible for standard image resolutions.

For the \textbf{general scene} setting, we collected 30,000 diverse images, including natural landscapes, classical paintings, and modern art. 
Since these images lack explicit style labels and each possesses a unique aesthetic, we form negative pairs by simply selecting two random images from the collection. 
The patch extraction and training process then follows the same protocol as the indoor scene setting.

\noindent\textbf{Optimization.} Our NSD framework is built upon the pre-trained Stable Diffusion v1.5 model and employs a two-stage training strategy. 
In the first stage, we train the PSRL module. To accommodate the distinct visual characteristics of different domains, we train separate PSRL modules for general and indoor scenes using publicly available and web-scraped data, which obviates the need for manual annotation. 
Each image is processed into ten 128$\times$128 patches, and the PSRL module is optimized using the loss function defined in Eq.~(\ref{eq:simple_loss_psrl}). 
In the second stage, the pre-trained PSRL module is frozen and integrated into the main NSD framework for end-to-end training. 
During this stage, only the newly introduced style-based cross-attention layers and the reference network are trainable. 
This partial fine-tuning approach is a deliberate design choice with a twofold rationale: 1) Freezing the core U-Net and text-attention layers of the base model preserves its powerful generative priors, mitigating the risk of catastrophic forgetting. 2) This strategy concentrates computational resources on learning stylistic coherence—the primary limitation of the original model—thereby ensuring an efficient and stable optimization process. 
For training, input images are resized so their shorter side is 512 pixels, then center-cropped to a final resolution of 512$\times$512. 
The trainable parameters are then updated via backpropagation using the diffusion loss from Eq.~(\ref{eq:simple_loss_NSD}).

\subsection{Quantitative Comparison}
We compare the NSD with competing baselines in the following three settings. As the first two settings, all models are trained on BrushData, and evaluated on BrushBench (see Tab. \ref{table:comparison_general}) to evaluate their general performance and evaluated on S3IMIndoorData to evaluate their generalization ability (see Tab. \ref{table:comparison_evalution_indoor}). For the third settings, all models are fine-tuned and tested on S3IMIndoorData to evaluated their performance on indoor scenes (see Tab. \ref{table:comparison_finetune_evaluation_indoor}). 

\noindent\textbf{Setting one.}
We begin by analyzing the quantitative results of models trained on BrushData and evaluated on BrushBench, summarized in Tab.~\ref{table:comparison_general}. 
\textbf{BLD}~\cite{avrahami2023blended} exhibits significant limitations as it fails to model scene style, leading to stylistically incongruent outputs reflected in low CSD, HPS, and IR scores. 
Furthermore, its latent blending mechanism struggles to create seamless transitions at mask boundaries. 
\textbf{SDI}~\cite{rombach2022high} employs an expanded U-Net, but its cross-attention mechanism's over-reliance on the text prompt compromises content preservation (evidenced by a low PSNR of 21.04) while simultaneously neglecting the intrinsic scene style. 
Although \textbf{BrushNet}~\cite{ju2024brushnet} and \textbf{CNI}~\cite{zhang2023adding} improve text alignment and image quality by conditioning on the masked image via dedicated branches, they still largely fail to capture and enforce stylistic consistency. 
\textbf{PowerPaint (PP)}~\cite{zhuang2023task} demonstrates improved stylistic adherence by interpolating task prompts; however, its learned style is often localized to the immediate context of the mask, failing to capture the global aesthetic of the entire scene. 
In contrast, our NSD framework leverages the PSRL module to extract a robust style representation and employs a dual-pathway mechanism to inject style (via cross-attention) and context (via the reference network), achieving state-of-the-art performance across all metrics.

\begin{table}[!t]
\centering
\resizebox{\columnwidth}{!}{
\begin{tabular}{ccccccc}
\toprule
Method & CSD $\uparrow$ & HPS$\uparrow$ & IR $\uparrow$ & Clip Sim $\uparrow$ & AS $\uparrow$ & PSNR$\uparrow$ \\
\midrule
BLD & 47.87& 25.81& 9.79 & 26.13 & 6.05& 20.93\\
SDI & 48.35& 26.11& 11.78 & 26.37 & 6.19& 21.04\\
PP & 48.54& 26.76& 11.98& 26.58 & 6.21 & 21.12\\
CNI & 48.23& 26.24& 11.87& 26.63 & 6.51 & 21.55 \\
BrushNet & 48.04& 26.42& 11.76& 26.61 & 6.54 & 21.54 \\
Ours & \textbf{49.18} & \textbf{27.25} & \textbf{12.06} & \textbf{26.63} & \textbf{6.55} & \textbf{21.57} \\
\bottomrule
\end{tabular}
}
\caption{Quantitative comparisons among NSD and other diffusion based models in BrushBench \cite{ju2024brushnet}.}
\label{table:comparison_general}
\end{table}

\noindent\textbf{Setting two.}
Indoor scenes typically exhibit stronger and more coherent stylistic patterns than general scenes. 
To assess the zero-shot generalization capabilities of models trained on BrushData, we evaluate them directly on our S3IMIndoorData test set. 
The results, presented in Tab.~\ref{table:comparison_evalution_indoor}, show that our method again achieves superior performance across all metrics, with a particularly notable advantage in stylistic consistency. 
This can be attributed to the design of our PSRL module. 
PSRL operates on the assumption that intra-scene patches share a common style while inter-scene patches do not—an assumption that holds especially true for indoor environments. 
Consequently, when applied to indoor scenes where style consistency is more pronounced, the PSRL module can extract a more precise and robust style representation. 
This, in turn, leads to a marked improvement in the stylistic harmony between the generated content and the surrounding scene. 
While \textbf{PowerPaint (PP)} also aims for consistency, its implicit approach, which lacks an explicit style representation mechanism, proves less effective than our targeted strategy.

\noindent\textbf{Setting three.}
To further investigate model performance in specialized domains, we fine-tuned our method, along with \textbf{CNI} and \textbf{BrushNet}, on our newly constructed S3IMIndoorData dataset. 
As shown in Tab.~\ref{table:comparison_finetune_evaluation_indoor}, fine-tuning on this domain-specific data improved the performance of all models across nearly every metric, validating the effectiveness of our dataset for adapting general-purpose models to indoor environments. 
Notably, our method maintained its superior performance, particularly in metrics related to stylistic consistency. 
This result further underscores the PSRL module's robust capability to extract and leverage scene-specific style representations, even in a fine-tuning scenario.

When interpreting these results, it is essential to consider our primary design goals. 
NSD was engineered specifically to enhance stylistic consistency and semantic alignment—objectives directly quantified by metrics like CSD, HPS, IR, and CLIP Score. 
Our consistent and significant improvements on these core metrics fundamentally validate the effectiveness of our approach. 
In contrast, PSNR and AS serve as crucial secondary metrics, evaluating background preservation and overall aesthetic quality, respectively. 
Therefore, our competitive performance on these metrics is a significant strength, demonstrating that our primary advancements in style and semantics are achieved without compromising background integrity or general visual appeal.

\begin{table}[!t]
\centering
\resizebox{\columnwidth}{!}{
\begin{tabular}{ccccccc}
\toprule
Model & CSD $\uparrow$ & HPS$\uparrow$ & IR $\uparrow$ & Clip Sim $\uparrow$ & AS $\uparrow$ & PSNR$\uparrow$ \\
\midrule
BLD & 38.12& 25.76& 3.01 & 12.45 & 5.98& 19.01\\
SDI & 38.77& 26.10& 4.18 & 12.51 & 6.12& 19.12\\
PP & 39.25& 26.82& 5.12& 12.62 & 6.23& 19.22\\
CNI & 38.79& 26.12& 4.48 & 12.54 & 6.35 & 19.98 \\
BrushNet & 38.69& 26.09& 4.81& 12.89 & 6.32& 19.92 \\
Ours & \textbf{40.70} & \textbf{27.19} & \textbf{5.28} & \textbf{13.22} & \textbf{6.37} & \textbf{20.01} \\
\bottomrule
\end{tabular}
}
\caption{Quantitative comparisons among NSD and other diffusion based models in the S3IMIndoorData without finetune in indoor scene.}
\label{table:comparison_evalution_indoor}
\end{table}

\begin{table}[!t]
\centering
\resizebox{\columnwidth}{!}{
\begin{tabular}{ccccccc}
\toprule
Model & CSD $\uparrow$ & HPS$\uparrow$ & IR $\uparrow$ & Clip Sim $\uparrow$ & AS $\uparrow$ & PSNR$\uparrow$ \\
\midrule
CNI & 39.88& 26.93& 4.92 & 13.58 & 6.39 & 19.95\\
Brushnet & 39.53& 27.29& 5.04& 13.86 & 6.33& 19.94\\
Ours & \textbf{41.82} & \textbf{28.16} & \textbf{5.34} & \textbf{14.02} & \textbf{6.48} & \textbf{20.12} \\
\bottomrule
\end{tabular}
}
\caption{Quantitative comparisons among NSD and other diffusion based models in the S3IMIndoorData after finetune.}
\label{table:comparison_finetune_evaluation_indoor}
\end{table}

\subsection{Qualitative Comparisons}
This section presents a qualitative comparison of the visual results, as illustrated in Fig.~\ref{fig:visual_baseline}. 
Both \textbf{CNI} and \textbf{BrushNet} succeed in generating semantically plausible content within the masked region. 
However, their failure to explicitly model scene style leads to a noticeable stylistic disconnect between the generated patch and its surrounding context, resulting in visually jarring and unnatural outcomes (see second and third columns). 
\textbf{PowerPaint (PP)} attempts style-aware manipulation by interpolating between "Context-aware Image Inpainting" and "Text-Guided Object Inpainting" task prompts. 
While this improves local consistency, its style perception is limited to the immediate vicinity of the mask. 
Consequently, for large regions with sparse contextual cues, such as the ceiling in the second row, it fails to generate realistic and coherent content. 
Furthermore, its unified network architecture, which simultaneously handles generation and preservation, appears to be overburdened, leading to severe artifacts. 
In stark contrast, our method leverages the PSRL module to capture a global style representation of the entire scene. 
By employing a dual-pathway architecture—using parallel cross-attention for style injection and a reference network for context preservation—our NSD framework robustly achieves results that are both semantically accurate and stylistically harmonious.

\begin{figure}[!t]
    \centering
    \includegraphics[width=0.48\textwidth]{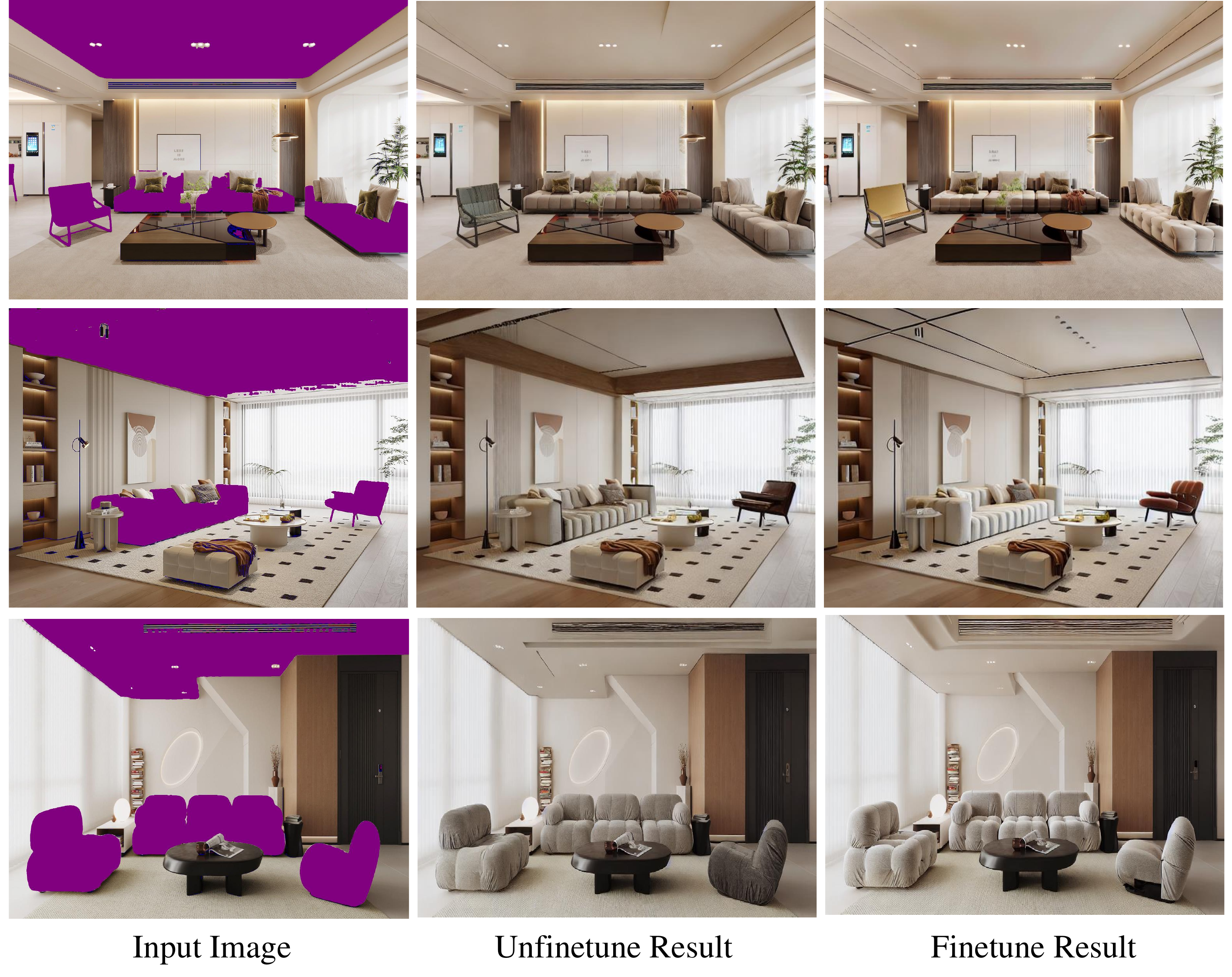}    
    \caption{Comparative results of NSD in finetune and unfinetune settings, please zoom in to see the details.}
    \label{fig:NSD_fine_unfine}
\end{figure}

We also visualize the impact of fine-tuning NSD on our S3IMIndoorData dataset in Fig.~\ref{fig:NSD_fine_unfine}, comparing the model's performance with and without this domain-specific adaptation. 
After fine-tuning, NSD achieves a superior level of stylistic integration between the generated content and the surrounding context, as evidenced by the results in the second row. 
This enhanced visual harmony underscores the effectiveness of our S3IMIndoorData dataset in enabling the model to learn and replicate the nuanced stylistic patterns specific to indoor environments.

\begin{figure}[!t]
    \centering
    \includegraphics[width=0.45\textwidth]{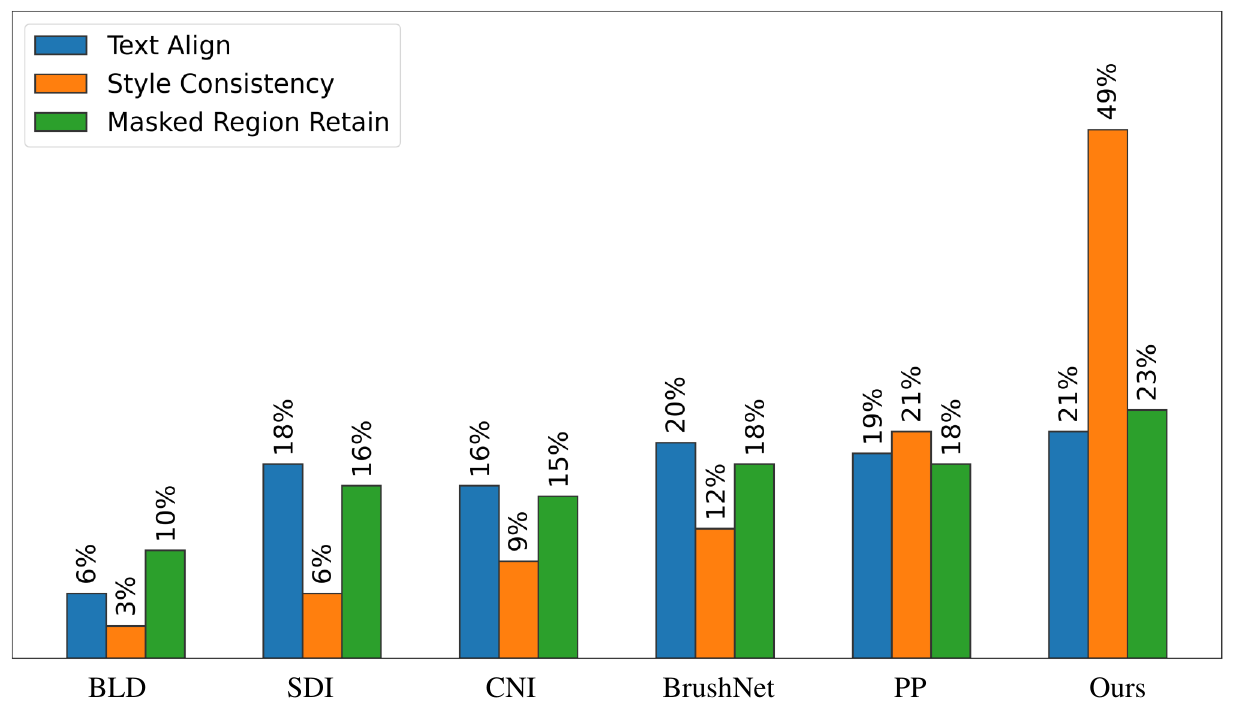}    
    \caption{Style consistency, Text alignment, and Masked region retention in the BrushBench user study across all models}
    \label{fig:userstudy}
\end{figure}

\begin{figure}[!t]
    \centering
    \includegraphics[width=0.45\textwidth]{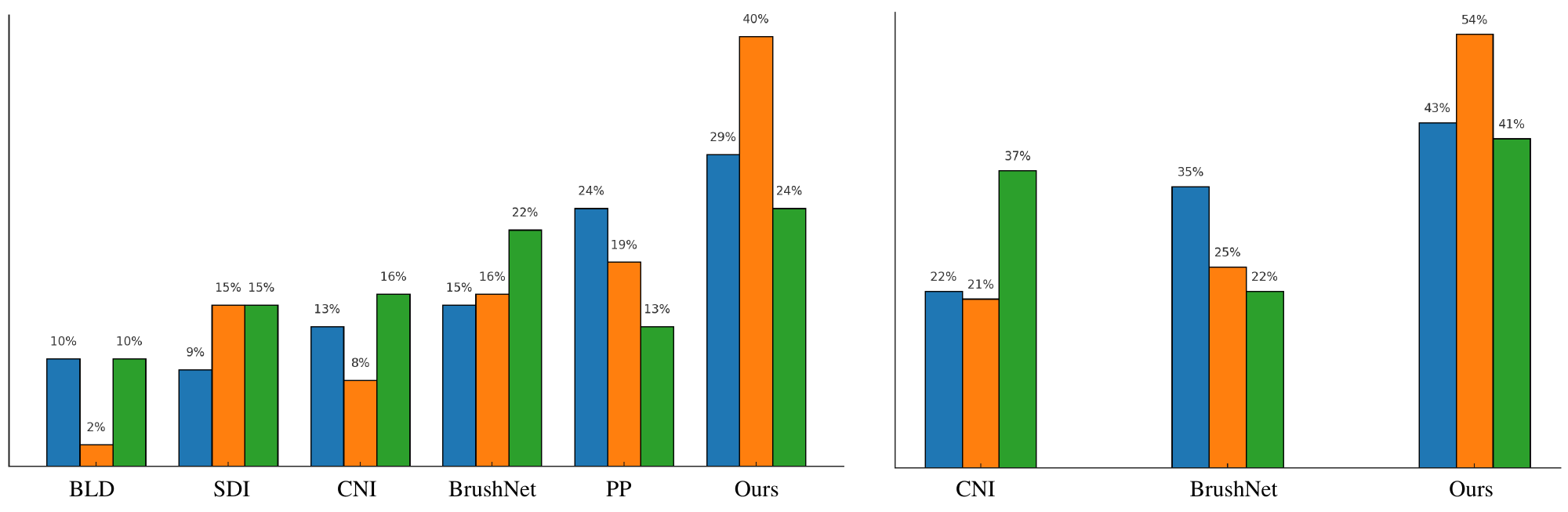}
    \vspace{-10pt}
    \caption{Style consistency, Text alignment, and Masked region retention user study in the second setting (left half: all models are trained on BrushData and evaluated on S3IMIndoorData), together with the corresponding user study in the third setting (right half: all models are fine-tuned and tested on S3IMIndoorData).}
    \label{fig:userstudy_plus}
\end{figure}

\subsection{User Study}
We recruited 25 users to evaluate 50 sets of generated images, assessing the results produced by all models from three perspectives: (1) style consistency between the generated and non-generated regions (\textbf{Style Consistency}), (2) alignment between the generated regions and semantic descriptions (\textbf{Text Align}), and (3) preservation of non-generated regions (\textbf{Masked Region Retain}). The user study was conducted under three different settings: first, all models were trained on \textbf{BrushData} and evaluated on BrushBench; second, the models were trained on BrushData but evaluated on S3IMIndoorData, testing their generalization ability across datasets; finally, all models were fine-tuned and tested on S3IMIndoorData to assess their performance in a domain-adapted scenario. As shown in Fig. \ref{fig:userstudy} and Fig. \ref{fig:userstudy_plus}, the proposed method was preferred by the majority of users, particularly in terms of style consistency.  

\subsection{Ablation Study}

\begin{figure}[!t]
    \centering
    \includegraphics[width=0.45\textwidth]{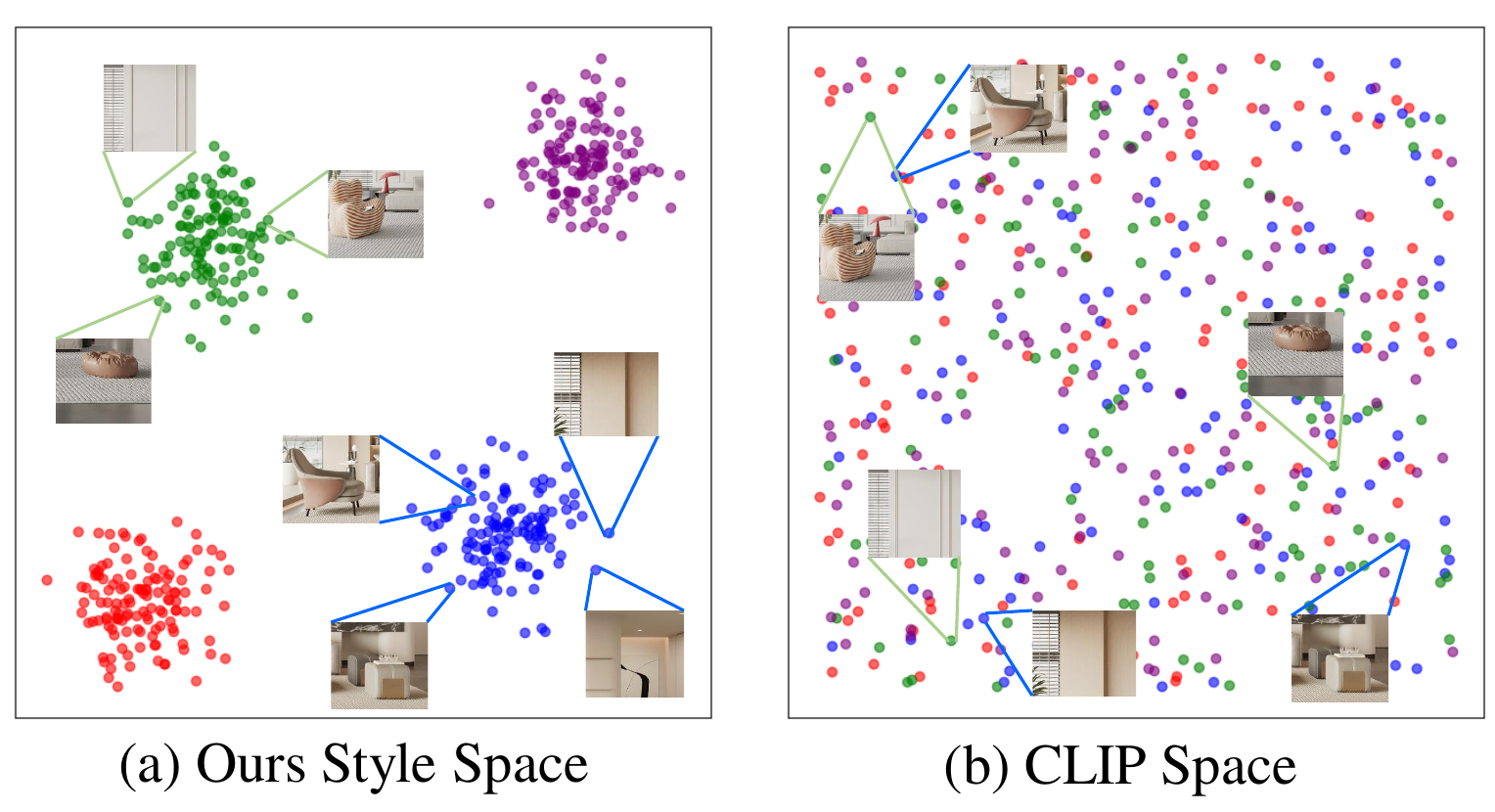}
    \caption{Although visual-language models like CLIP \cite{radford2021learning} have demonstrated remarkable capability in extracting features from open-domain images, it struggle to capture style representation in scenes. Our PSRL module excels in this aspect.}
    \label{fig:motivation}
\end{figure}

\noindent\textbf{Analyses of PSRL .}
We emphasize in the introduction that Vision-Language Models (VLMs) are ill-suited for accurately extracting stylistic information from a scene. 
To validate this assertion, we replace the PSRL module with a style encoder using the CLIP image encoder and a learnable adapter (denoted as "Ours w/ CLIP Style"). 
As shown in Tab.~\ref{table:comparison_ablation_psrl}, this "Ours w/ CLIP Style" variant leads to a significant performance drop on metrics measuring stylistic consistency, namely CSD, HPS, and IR. 
This result demonstrates that a VLM backbone like the CLIP image encoder is inadequate for capturing salient stylistic cues in scenes. 
To further analyze this phenomenon, we performed a dimensionality reduction and visualization of the style representations extracted by PSRL versus those from the VLM's image encoder. 
As depicted in Fig.~\ref{fig:motivation}, the VLM struggles to capture style cues, as its extracted features are predominantly organized by semantic information. 
In contrast, the style representations learned by PSRL effectively form distinct clusters of patches with the same style while separating those with different styles. 
This provides strong validation that PSRL can successfully disregard semantic information and learn accurate style representations.

\begin{table}[!t]
\centering
\resizebox{\columnwidth}{!}{
\begin{tabular}{ccccccc}
\toprule
Method & CSD $\uparrow$ & HPS$\uparrow$ & IR $\uparrow$ & Clip Sim $\uparrow$ & AS $\uparrow$ & PSNR$\uparrow$ \\
\midrule
Ours base CLIP & 39.91 & 27.62 & 5.13 & 14.03 & 6.51 & 20.11 \\
Ours base category & 40.79 & 27.75 & 5.09 & 14.01 & 6.50 & 20.10 \\ 
Ours based statistics & 41.01 & 27.84 & 5.12 & 13.99 & 6.47 & 20.13 \\
Ours w/o Progressive & 39.73 & 26.96 & 5.01 & 14.06 & 6.46 & 20.11 \\
Ours & \textbf{41.82} & \textbf{28.16} & \textbf{5.34} & \textbf{14.02} & \textbf{6.48} & \textbf{20.12} \\
\bottomrule
\end{tabular}
}
\caption{Ablation experiments on the PSRL module in the finetune setting}
\label{table:comparison_ablation_psrl}
\end{table}

To validate the necessity of the full progressive pipeline, we first evaluate a simplified approach that relies solely on second-order statistics (``Ours based on statistics"). As shown in Tab. \ref{table:comparison_ablation_psrl}, this method provides a stable baseline but ultimately yields suboptimal stylistic consistency. The theoretical reason is that while these statistics effectively summarize global style properties in a manner robust to local content variations, they are too coarse to capture the fine-grained textures and nuanced details that define a unique style. The network learns to fit the statistical distribution rather than the style itself.

\begin{figure}[!t]
    \centering
    \includegraphics[width=0.45\textwidth]{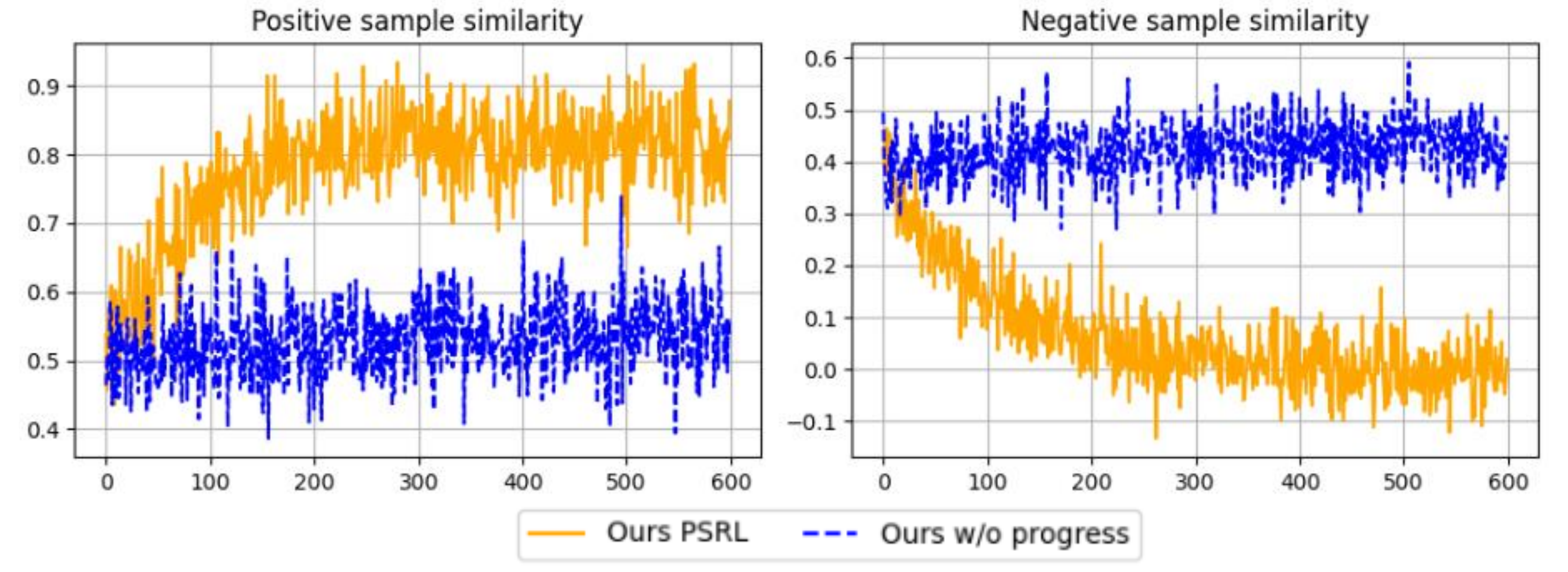}
    \caption{The convergence process of positive and negative sample similarity during training for Ours PSRL and Ours w/o Progressive.}
    \label{fig: ablation psrl}
\end{figure}

Next, we analyze the effect of applying contrastive learning directly, without the statistical pre-training stage (``Ours w/o Progressive"). This approach struggles significantly, failing to improve stylistic consistency and showing poor convergence (Fig. \ref{fig: ablation psrl}). This failure stems from the high semantic disparity often found in complex scenes (\textit{e.g.}, patches of 'sky' vs. 'tree'). Forcing a model to treat such semantically distinct patches as stylistically identical creates a high-variance, unstable learning signal with conflicting gradients. Our progressive strategy directly resolves this issue. We first use the robust, low-variance signal from second-order statistics to pre-train a foundational backbone that is sensitive to global style but resilient to local semantic conflicts. Upon this stable foundation, the subsequent contrastive learning stage can effectively and stably refine the representation to capture fine-grained style details.

In the introduction, we argue that learning style representations from coarse category-level supervision (e.g., "minimalist" or "European"), as done in~\cite{zhang2022domain, yang2024emogen}, fails to capture the nuanced stylistic variations unique to each image. 
To empirically investigate this claim, we conduct an ablation study. 
We first train an indoor scene classifier with category-level supervision and then use the features from its final layer as a style representation, integrating this into our NSD framework (denoted as "Ours w/ Category Style"), mirroring the approach in~\cite{yang2024emogen}. 
As shown in Tab.~\ref{table:comparison_ablation_psrl}, this variant's performance on stylistic consistency metrics is notably inferior to that of PSRL, which operates at the instance level. 
Unlike representations learned from broad style categories, PSRL's instance-level objective forces the model to find a consistent representation across semantically diverse patches from the \textit{same} image. 
This process inherently encourages the model to disregard variable semantic content and focus on the underlying, unifying style. 
Consequently, PSRL can discern subtle stylistic differences between images while ensuring strong stylistic consistency within a single image, yielding a more fine-grained and effective style representation.

\noindent\textbf{Analyses of Parallel Mechanisms.}
We employ two parallel cross-attention mechanisms to independently process textual and stylistic features. 
To validate this design choice, we compare it against a baseline that instead concatenates the text and style features before feeding them into a single cross-attention layer. 
As shown in Tab.~\ref{table:comparison_ablation_dca}, this concatenation approach leads to a discernible decline in both image quality and stylistic consistency. 
A plausible explanation is that the key and value projection matrices within the pre-trained diffusion model's cross-attention layers are specifically optimized for text-based feature distributions. 
Fusing style features directly with text features before this projection can disrupt the expected input manifold, causing the model to neglect fine-grained, image-specific style information and resulting in only coarse-grained stylistic control.

\begin{table}[!t]
\centering
\resizebox{\columnwidth}{!}{
\begin{tabular}{ccccccc}
\toprule
Method & CSD $\uparrow$ & HPS$\uparrow$ & IR $\uparrow$ & Clip Sim $\uparrow$ & AS $\uparrow$ & PSNR$\uparrow$ \\
\midrule
Ours w/o parallel & 40.73 & 27.88 & 5.16 & 13.81 & 6.21 & 20.09 \\
Ours & \textbf{41.82} & \textbf{28.16} & \textbf{5.34} & \textbf{14.02} & \textbf{6.48} & \textbf{20.12} \\
\bottomrule
\end{tabular}
}

\caption{Ablation experiments on the two parallel cross-attention mechanisms in the finetune setting}
\label{table:comparison_ablation_dca}
\end{table}

\section{Limitations and Future Work}
\label{sec:limitations}
Our PSRL module's core assumption of a single, globally consistent style per image, while effective for stylistically coherent domains like indoor and many general scenes, presents limitations in more diverse contexts. Images with intentionally mixed aesthetics (\textit{e.g.}, art collages) can challenge our global sampling strategy, potentially causing it to learn a non-representative ``average" style. This may lead to edits that fail to harmonize with any specific local aesthetic. A compelling direction for future work is to develop more adaptive, locally-aware style extraction mechanisms. Such models could learn to disentangle multiple style regions or incorporate user guidance for style referencing, thereby broadening our framework's applicability to more complex visual content.

\section{Conclusion}
In this work, we introduce the Neural Scene Designer (NSD), a novel framework that leverages an advanced diffusion model and incorporates two parallel cross-attention mechanisms to seamlessly integrate textual and stylistic information. Additionally, a reference network aggregates features from unmasked regions and incorporates them into the diffusion denoising process, ensuring content coherence throughout the image. We also present the Progressive Self-style Representational Learning (PSRL) module, which captures fine-grained style representations by leveraging the idea that regions within a scene share a consistent style, while regions across different scenes have distinct styles. NSD ensures semantic alignment with user intent and stylistic consistency with the environment. Extensive evaluations on novel and publicly available datasets validate its effectiveness.

{
    \small
    \bibliographystyle{ieeenat_fullname}
    \bibliography{main}
}

\begin{IEEEbiography}[{\includegraphics[width=1in,height=1.25in,clip,keepaspectratio]{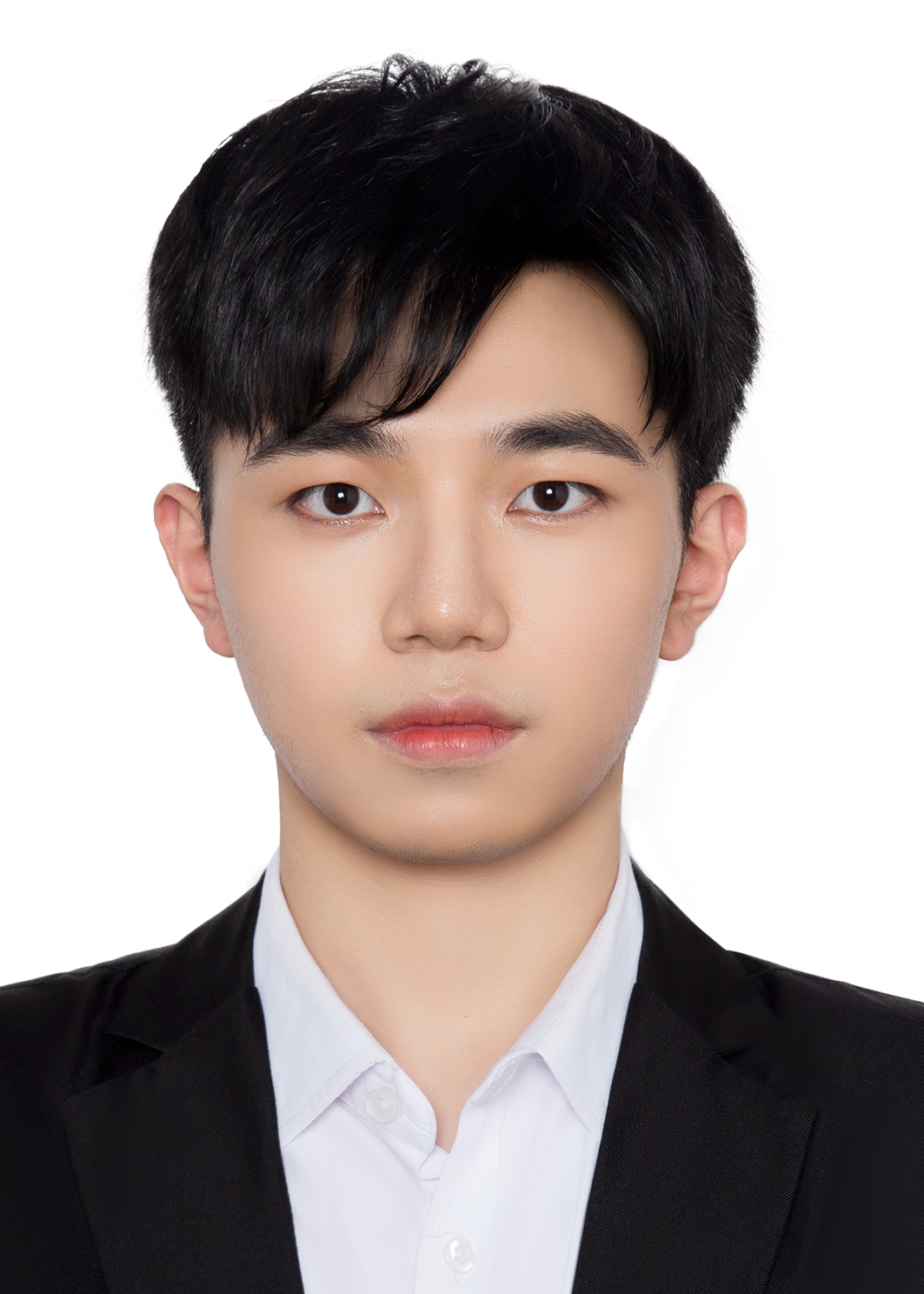}}]{Jianman Lin} received a B.Sc. degree from Guangdong University of Technology, Guangzhou, China, in 2024. He is currently pursuing a Master's degree in Electronic Information at South China University of Technology. His research interests include artificial intelligence, multimodal large models, and generative AI. He has published papers in prestigious conferences and journals, including CVPR and IJCV.\end{IEEEbiography}

\begin{IEEEbiography}[{\includegraphics[width=1in,height=1.25in,clip,keepaspectratio]{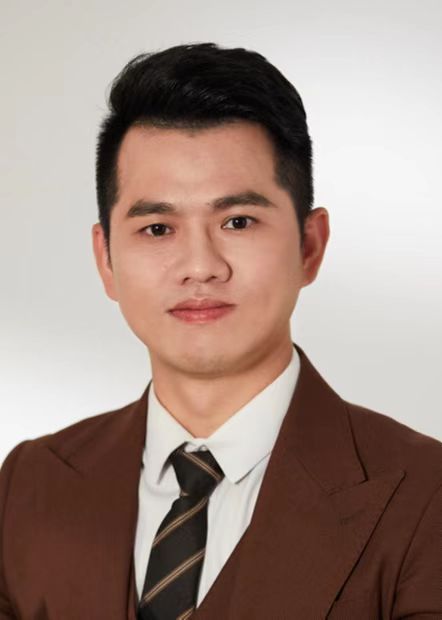}}]{Tianshui Chen} received a Ph.D. degree in computer science at the School of Data and Computer Science Sun Yat-sen University, Guangzhou, China, in 2018. Prior to earning his Ph.D, he received a B.E. degree from the School of Information and Science Technology in 2013. He is currently an associate professor at the Guangdong University of Technology. His current research interests include artificial intelligence, multimodal large models, and generative AI. He has authored and co-authored more than 60 papers published in top-tier academic journals and conferences, including T-PAMI, IJCV, T-NNLS, T-IP, T-MM, CVPR, ICCV, AAAI, IJCAI, ACM MM, etc. He has served as a reviewer for numerous academic journals and conferences. He was the recipient of the Best Paper Diamond Award at IEEE ICME 2017. \end{IEEEbiography}

\begin{IEEEbiography}[{\includegraphics[width=1in,height=1.25in,clip,keepaspectratio]{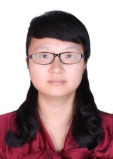}}]{Chunmei Qing}  received her B.Sc. degree in Information and Computation Science from Sun Yat-sen University, China, in 2003, and Ph.D. degree in Electronic Imaging and Media Communications from University of Bradford, UK, in 2009. Then she worked as a postdoctoral researcher in the University of Lincoln, UK. From 2013 till now, she is an associate professor in School of Electronic and Information Engineering, South China University of Technology (SCUT), Guangzhou, China. Her main research interests include image/video processing, computer vision, and affective computing.
\end{IEEEbiography}

\begin{IEEEbiography}[{\includegraphics[width=1in,height=1.25in,clip,keepaspectratio]{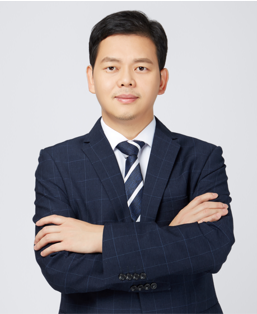}}]{Zhijing Yang} received the B.S and Ph.D. degrees from the Mathematics and Computing Science, Sun Yat-sen University, Guangzhou China, in 2003 and 2008, respectively. He was a Visiting Research Scholar in the School of Computing, Informatics and Media, University of Bradford, U.K, between July-Dec, 2009, and a Research Fellow in the School of Engineering, University of Lincoln, U.K, between Jan. 2011 to Jan. 2013. He is currently a Professor and Vice Dean at the School of Information Engineering, Guangdong University of Technology, China. He has published over 80 peer-reviewed journal and conference papers, including IEEE T-CSVT, T-MM, T-GRS, PR, etc. His research interests include machine learning and pattern recognition.
\end{IEEEbiography}

\begin{IEEEbiography}[{\includegraphics[width=1in,height=1.25in,clip,keepaspectratio]{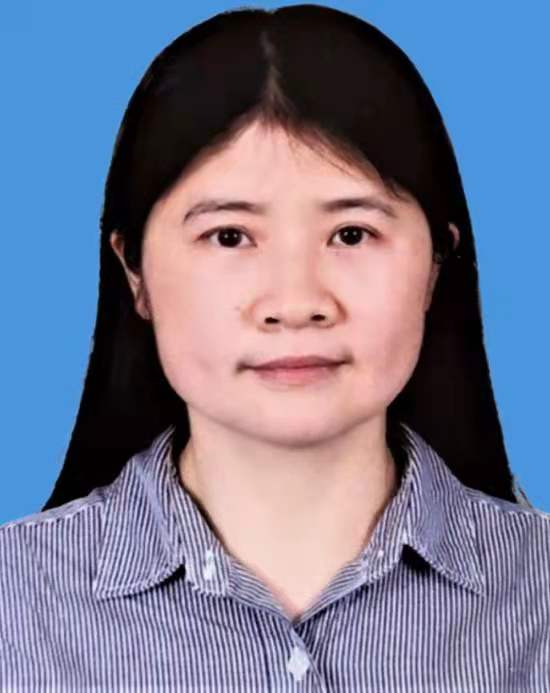}}]
{Shuangping Huang} received the PhD degree from the South China University of Technology in 2011. She is currently a professor with the school of electronic and information engineering, the South China University of Technology. Her current research interests are embodied intelligence, AIGC, explainable and trustworthy artificial intelligence, text intelligence, computer vision, and deep learning. 
\end{IEEEbiography}

\begin{IEEEbiography}[{\includegraphics[width=1in,height=1.25in,clip,keepaspectratio]{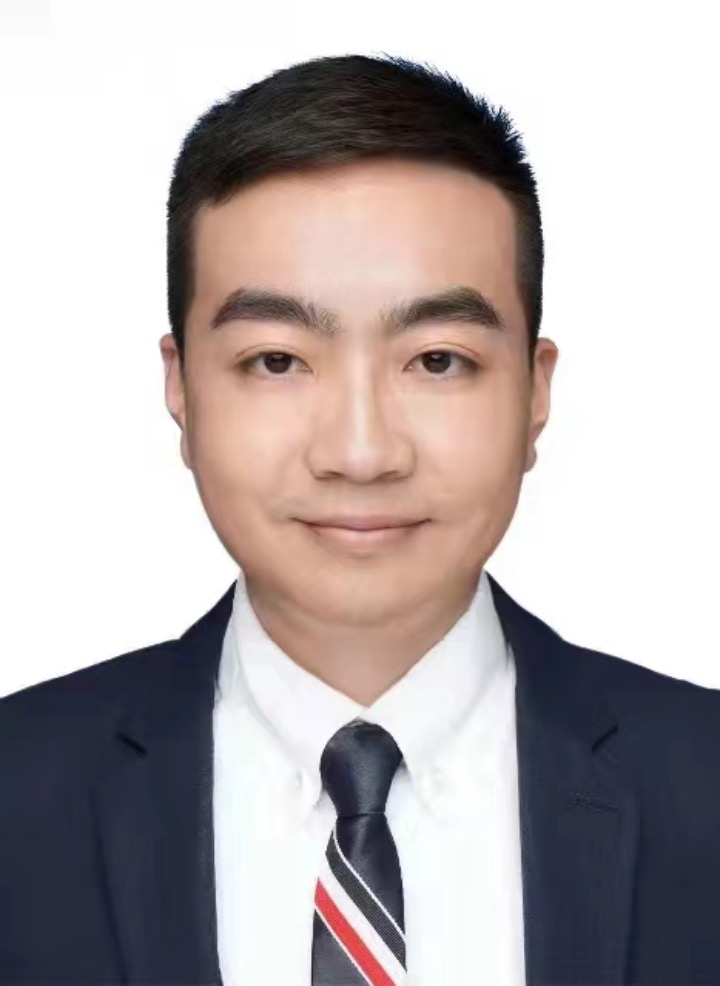}}]
{Yuheng Ren} holds a PhD of Information Management  and a PhD of Information Economics . He is an IET Fellow and an RSA Life Fellow. Currently, he serves as the Executive Dean of the Digital Industry College at Jimei University and the Director of the Xiamen Kunlu AI Research Institute. He has published 17 monographs, obtained 33 patents, and authored 49 high-level papers. In 2023, he was awarded the EU Research and Innovation Award. In 2024, he received the First Prize for Technological Progress from the China General Chamber of Commerce and the Second Prize for Technological Progress from the China Federation of Logistics and Purchasing. 
\end{IEEEbiography}

\begin{IEEEbiography}[{\includegraphics[width=1in,clip]{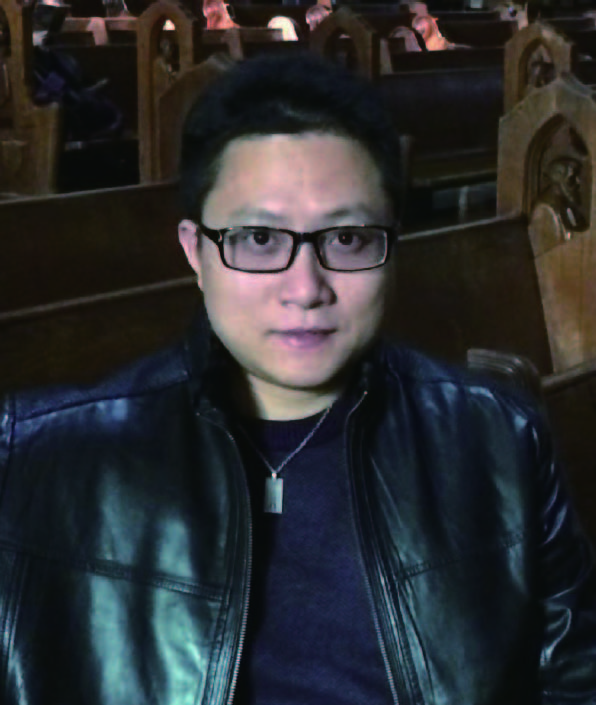}}]{Liang Lin} (Fellow, IEEE) is a full professor at Sun Yat-sen University. From 2008 to 2010, he was a postdoctoral fellow at the University of California, Los Angeles. From 2016--2018, he led the SenseTime R\&D teams to develop cutting-edge and deliverable solutions for computer vision, data analysis and mining, and intelligent robotic systems. He has authored and co-authored more than 100 papers in top-tier academic journals and conferences (e.g., 15 papers in TPAMI and IJCV and 60+ papers in CVPR, ICCV, NIPS, and IJCAI). He has served as an associate editor of IEEE Trans. Human-Machine Systems, The Visual Computer, and Neurocomputing and as an area/session chair for numerous conferences, such as CVPR, ICME, ACCV, and ICMR. He was the recipient of the Annual Best Paper Award by Pattern Recognition (Elsevier) in 2018, the Best Paper Diamond Award at IEEE ICME 2017, the Best Paper Runner-Up Award at ACM NPAR 2010, Google Faculty Award in 2012, the Best Student Paper Award at IEEE ICME 2014, and the Hong Kong Scholars Award in 2014. He is a Fellow of IEEE, IAPR, and IET. \end{IEEEbiography}

\end{document}